\documentclass[11pt]{article}

\usepackage[preprint]{acl}

\usepackage{times}
\usepackage{latexsym}

\usepackage[T1]{fontenc}
\usepackage[utf8]{inputenc}

\usepackage{microtype}

\usepackage{inconsolata}

\usepackage{graphicx}
\usepackage{booktabs}
\usepackage{multirow}
\usepackage[table]{xcolor}
\usepackage{amsmath, amssymb}
\usepackage{float}
\usepackage{algorithm}
\usepackage{algpseudocode}

\title{Rethinking AI-Generated Text Detection: A Strong Baseline and the Distribution-Shift Problem That Remains}

\author{
Zhuoer Shen$^1$ \quad
Mingyi Wang$^2$ \quad
Shaofeng Zou$^2$ \quad
Yuheng Bu$^1$\thanks{Corresponding author.} \\
$^1$Department of Computer Science, University of California, Santa Barbara \\
$^2$School of ECEE, Arizona State University \\
\texttt{zhuoershen@ucsb.edu, mwang287@asu.edu} \\
\texttt{zou@asu.edu, buyuheng@ucsb.edu}
}

\begin{document}
\maketitle

\begin{abstract}
Recent AI-generated text detection work often introduces a new benchmark together with a specialized detector tailored to it. We revisit this practice from a baseline-first perspective. Across several benchmarks, we show that a plain, fully fine-tuned RoBERTa matches or exceeds the specialized detectors those benchmarks are built around. This suggests that much of the recent architectural complexity is not what drives strong in-distribution detection. The remaining challenge is the distribution shift. The same strong baseline degrades sharply when the topic domain or generating model changes at test time, and simply adding more source data does not close the gap. We identify a key failure mode: under distribution shift, the detector can assign high-confidence machine labels to human-written text from unseen domains. We then study two lightweight domain adaptation methods to address this problem: $K$-shot adaptation with first-order MAML over LoRA adapters, and a per-sample confidence-weighted ensemble built on top of the adapted detector. Overall, our results suggest that progress in AI-generated text detection should be measured not only by in-distribution performance, but also by robustness under distribution shift.

\end{abstract}

\section{Introduction}
\label{sec:intro}
Large language models (LLMs) now generate text that is difficult to distinguish from human writing, raising concerns in settings such as education~\citep{susnjak2022chatgpt}, peer review~\citep{yu2025peerreview}, and the integrity of online information~\citep{spitale2023ai}. This has motivated a large and fast-growing body of work on AI-generated text detection.

Recent AI-generated text detection work has continued to introduce new detectors alongside new benchmarks. In several cases, the benchmark is released together with a \emph{specialized} detector reported to outperform prior methods on that benchmark: the IntelLabs peer-review benchmark releases Anchor~\citep{yu2025peerreview}, calibrated for strict low-FPR operation; FAID couples a three-way detector with a vector-database augmentation mechanism~\citep{ta2026faid}; and MIRAGE proposes DetectAnyLLM~\citep{fu2025detectanyllm}. A related line of work builds large ``in-the-wild'' benchmarks and reports which existing detector performs best on them; MAGE~\citep{li2024mage}, for instance, finds a long-context pre-trained language model detector to be the strongest among the methods it evaluates. 

A striking commonality across these works is the baseline they improve upon: the GPT-2 classifier released by \citet{solaiman2019release}, used in a \emph{zero-shot}, un-fine-tuned manner. Such a baseline is useful as an early off-the-shelf reference, but it does not answer the more basic question: how strong is a standard supervised baseline trained on the same data? A zero-shot detector and a classifier fine-tuned on the benchmark’s own training set are fundamentally different points of comparison. Without evaluating the latter under the same protocol, it remains unclear whether the reported gains come from detector design or from access to in-distribution training data.

\begin{figure*}[t]
\centering
\includegraphics[width=0.92\textwidth]{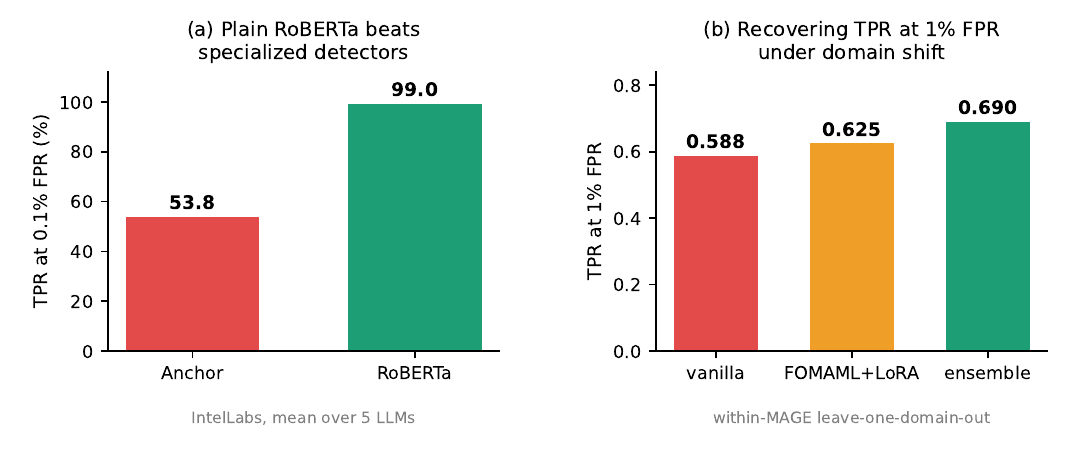}
\vspace{-1em}
\caption{Overview. (a) On IntelLabs peer-review benchmark~\citep{yu2025peerreview}, a fully fine-tuned RoBERTa baseline substantially outperforms the specialized Anchor detector over five source LLMs. (b) Under within-MAGE~\citep{li2024mage} leave-one-domain-out evaluation, two simple approaches to address distribution shift (FOMAML$+$LoRA and a confidence-weighted ensemble) both partially recover the TPR at $1\%$ FPR that the vanilla detector loses. The two panels use different benchmarks and FPR thresholds and are not directly comparable.}
\label{fig:overview}
\end{figure*}

We revisit the problem from this baseline-first perspective in two parts. \textbf{First, a strong generic baseline already goes surprisingly far.} We fine-tune RoBERTa~\citep{liu2019roberta} end-to-end on four detection benchmarks and find that it matches or exceeds the detector each benchmark is built around. As shown in Figure~\ref{fig:overview} (a), on IntelLabs, a plain fully fine-tuned RoBERTa substantially exceeds the specialized Anchor detector in TPR at $0.1\%$ FPR across five source LLMs. On MAGE, where the strongest reported detector uses a long-context window, a RoBERTa baseline restricted to $512$ tokens matches it, indicating that long-context modeling is not necessary to reach the reported detection accuracy.

\textbf{Second, distribution shift remains the central challenge.} A fully fine-tuned detector can be very strong when the test distribution is well covered by the training distribution. In deployment, however, no training set can enumerate every topic domain and every generator a detector may encounter, making robustness to \emph{distribution shift} the more consequential question. We find that a detector with near-perfect in-distribution scores degrades sharply when the topic domain or the generating model changes, and identify a specific failure mode behind this degradation: the detector confidently mislabels a majority of human-written text as machine, a failure that AUROC can obscure but low-FPR evaluation exposes. We then show that two simple approaches partially address this failure mode (Figure~\ref{fig:overview}(b)): a $K$-shot domain adaptation method using first-order Model-Agnostic Meta-Learning over Low-Rank adapters (FOMAML+LoRA)~\citep{nichol2018firstorder, hu2022lora}, and a per-sample confidence-weighted ensemble of the adapted and vanilla detectors.

% \textbf{Second, the same baseline breaks under distribution shift.} A fully fine-tuned detector can be very strong when the test distribution is well covered by the training distribution. In deployment, however, no training set can enumerate every topic domain and every generator a detector may encounter, making robustness to \emph{distribution shift} the more consequential question. We find that a detector with near-perfect in-distribution scores degrades sharply when the topic domain or the generating model changes. We address this shift on the target side, rather than by simply accumulating more source data, through two methods: $K$-shot first-order Model-Agnostic Meta-Learning (MAML) over Low-Rank adapters (LoRA)~\citep{nichol2018firstorder, hu2022lora}, and a confidence-weighted ensemble of the adapted and original detectors. The gains from these methods are illustrated in Figure~\ref{fig:overview}(b) over MAGE.

% The ensemble improves true-positive rates at low false-positive rates by reducing a high-confidence false-positive tail that neither detector removes on its own.

We summarize our contributions as follows: (1)~we show that a fully fine-tuned RoBERTa baseline matches or exceeds the specialized detectors proposed alongside three benchmarks (IntelLabs, FAID, and MIRAGE), and also matches the strongest existing detector reported on the in-the-wild MAGE benchmark once input length is held fixed. This suggests that progress in AI-generated text detection should be evaluated against strong fine-tuned baselines, rather than zero-shot detectors. (2)~we identify a key failure mode under distribution shift: strong fine-tuned detectors can assign high-confidence machine labels to human-written text from unseen domains. Low-FPR evaluation exposes this failure, and $K$-shot adaptation with a confidence-weighted ensemble can partially correct it.

% We summarize our contributions: (1)~we show that a fully fine-tuned RoBERTa baseline matches or exceeds the specialized detectors proposed alongside three benchmarks (IntelLabs, FAID, and MIRAGE), and also matches the strongest existing detector reported on the in-the-wild MAGE benchmark once input length is held fixed. This suggests that progress in AI-generated text detection should be evaluated against strong fine-tuned baselines, rather than zero-shot detectors. (2)~we characterize a distribution-shift gap that this strong baseline does not close, and identify a high-confidence false-positive tail as a key failure mode, showing through both $K$-shot adaptation and a simple confidence-weighted ensemble that this tail is reachable by lightweight target-side methods.

% We summarize our contributions: (1)~we show that a fully fine-tuned RoBERTa baseline matches or exceeds the specialized detectors proposed alongside three detection benchmarks, and matches the strongest reported detector on MAGE once input length is held fixed. This suggests that progress in AI-generated text detection should be evaluated against strong fine-tuned baselines, rather than zero-shot detectors. (2)~we characterize the distribution-shift problem that remains and study two target-side responses, including a confidence-weighted ensemble used as a simple probe into a high-confidence false-positive tail, demonstrating that this failure mode is reachable even by lightweight means.

\begin{table*}[t!]
\centering
\caption{The four AI-generated text detection benchmarks we considered. ``Generators'' lists either individual LLMs or LLM families (each family covering multiple specific models). Train and Test sizes are in number of examples; Test reports the main in-distribution test split, with additional evaluation splits described in Appendix~\ref{app:datasets}.}
\label{tab:datasets}
\small
\setlength{\tabcolsep}{4pt}
\begin{tabular}{lccccc}
\toprule
Dataset & Train & Test & Label & Domains & Generators \\
\midrule
IntelLabs~\citep{yu2025peerreview}     & 23K   & 192K  & binary & 1 (peer reviews) & 5 LLMs \\
MAGE~\citep{li2024mage}                & 319K  & 61K   & binary & 10 (mixed)       & 27 LLMs \\
FAID~\citep{ta2026faid}                & 61K   & 13K   & 3-way  & 2 (academic)     & 4 families \\
MIRAGE~\citep{fu2025detectanyllm}      & 1K$^{\dagger}$ & 188K  & binary & 5 (mixed) & 17 LLMs \\
\bottomrule
\end{tabular}
\par\raggedright\footnotesize $^{\dagger}$MIRAGE's training pool consists of 500 paired (human-machine) examples, i.e., $1{,}000$ samples total.
\end{table*}

\section{Related Work}
\label{sec:related}

\vspace{0.2em}\noindent\textbf{AI-generated text detection.}
AI-generated text detection spans zero-shot statistical methods and supervised neural detectors. Zero-shot approaches detect machine-generated text by exploiting likelihood artifacts, curvature, conditional probability, or disagreement patterns~\citep{gehrmann2019gltr, solaiman2019release, mitchell2023detectgpt, bao2024fastdetectgpt, hans2024binoculars}. Supervised and hybrid systems introduce richer features, adversarial training, fine-grained labels, or multi-domain benchmarks~\citep{verma2024ghostbuster, hu2023radar, li2024mage, abassy2024detectaive, yu2025peerreview, ta2026faid, fu2025detectanyllm}. However, this progress has often been measured against fixed or weakly adapted baselines from zero-shot or off-the-shelf settings. Although fine-tuned RoBERTa-style classifiers have appeared in prior studies, they are rarely treated as the main point of comparison under matched data, evaluation protocols, and deployment-relevant operating conditions. In contrast, our work takes a baseline-first perspective: we re-evaluate a fully fine-tuned supervised baseline and measure how much progress remains when the baseline has comparable training access and is evaluated at strict false-positive-rate thresholds.

% \begin{table*}[t!]
% \centering
% \caption{The five AI-generated text detection benchmarks considered in this work. ``Generators'' lists either individual LLMs or LLM families (each family covering multiple specific models).}
% \label{tab:datasets}
% \small
% \setlength{\tabcolsep}{4pt}
% \begin{tabular}{lcccc}
% \toprule
% Dataset & Size & Label & Domains & Generators \\
% \midrule
% IntelLabs~\citep{yu2025peerreview}     & 789K & binary & 1 (peer reviews)        & 5 LLMs \\
% MAGE~\citep{li2024mage}          & 437K & binary & 10 (mixed)              & 27 LLMs \\
% FAID~\citep{ta2026faid}          & 83K  & 3-way  & 2 (academic)            & 4 families \\
% MIRAGE~\citep{fu2025detectanyllm} & 500$^{\dagger}$ & binary & 5 (mixed)   & 17 LLMs \\
% HC3~\citep{guo2023close}         & 24K  & binary & 5 (finance, med., etc.) & ChatGPT \\
% \bottomrule
% \end{tabular}
% \par\raggedright\footnotesize $^{\dagger}$MIRAGE's training pool consists of 500 paired (human-machine) examples; its full test set spans 6 task subsets across 5 text-domains.
% \end{table*}

\vspace{0.2em}
\noindent\textbf{Distribution shift, adaptation, and deployment-oriented evaluation.}
A second line of work concerns the robustness of detectors under distribution shift. Distribution shift is a well-studied problem in machine learning~\citep{koh2021wilds, wang2022domaingeneralization}, and is especially acute for AI-generated text detection, where the target distribution changes with writing domain, prompt style, paraphrasing, and the underlying LLM. Existing detection benchmarks and recent domain-generalization studies explicitly show that detectors can perform well in-domain but fail to generalize to unseen generators or domains~\citep{li2024mage, borile2025generalize, eagle2024domain}. We address this setting through domain adaptation, using FOMAML over LoRA adapters~\citep{finn2017maml, nichol2018firstorder, hu2022lora, houlsby2019adapters, li2021prefixtuning}, which combines gradient-based meta-learning with parameter-efficient fine-tuning. Beyond adaptation, we use a per-sample confidence-weighted combination of the adapted and vanilla detectors and evaluate it under low-FPR operating points, where the failure mode of interest becomes visible.

% \paragraph{Domain shift, adaptation, and deployment-oriented evaluation.}
% A second line of work concerns the robustness of detectors under distribution shift. Distribution shift is a well-studied problem in machine learning~\citep{koh2021wilds, wang2022domaingeneralization}, and is especially acute for AI-generated text detection, where the target distribution changes with writing domain, prompt style, paraphrasing, and the underlying LLM. Existing detection benchmarks and recent domain-generalization studies explicitly show that detectors can perform well in-domain but fail to generalize to unseen generators or domains~\citep{li2024mage, borile2025generalize, eagle2024domain}. To address this setting, our work brings together two mature adaptation schemes that have not been systematically applied to AI-generated text detection: gradient-based meta-learning and parameter-efficient adaptation~\citep{finn2017maml, nichol2018firstorder, hu2022lora, houlsby2019adapters, li2021prefixtuning}. 

% We use this perspective to study target-side adaptation for AI-generated text detection, focusing not only on aggregate AUROC but also on low-FPR operating points such as TPR@1\%FPR, which better reflect deployment settings where false accusations of human authorship are costly.

\section{A Strong Fully Fine-Tuned In-Distribution Baseline}
\label{sec:baseline}

\subsection{Problem Setup}
\label{ssec:problem}

We study AI-generated text detection as a binary classification problem: given a piece of input text $x$, predict whether it is human-written ($y=0$) or machine-generated ($y=1$). A detector is a function $f: \mathcal{X} \to [0, 1]$ producing a probability $f(x)$ that $x$ is machine-generated. FAID~\citep{ta2026faid} instead uses a three-way label (human/machine/mixed); we train a three-way variant for that comparison as detailed in Appendix~\ref{app:datasets}.

This section considers the \emph{in-distribution} setting, where the detector is trained and evaluated on data from the same benchmark. Specifically, each detector is trained on the benchmark’s training split and evaluated on its held-out test split, following the benchmark’s standard protocol. Our goal is to assess how a standard fully fine-tuned classifier performs under the same conditions in which specialized detectors are typically evaluated.

% The complementary case, in which training and test distributions diverge, is the subject of Section~\ref{sec:shift}.

% \medskip
% \noindent

We evaluate the baseline on four publicly available AI-generated text detection benchmarks (Table~\ref{tab:datasets}), covering peer reviews, academic writing, and diverse open-domain genres, with machine-generated text produced by a broad set of LLM families; the same four benchmarks are used throughout the paper, and full details are given in Appendix~\ref{app:datasets}. As evaluation metrics, we report AUROC, a threshold-independent summary, together with the true-positive rate at a low, fixed false-positive rate (TPR at $1\%$ FPR, and $0.1\%$ FPR), when a stricter operating point is informative. The low-FPR metric is more relevant for deployment, where detectors use fixed thresholds in practice and falsely flagging human-written text is costly.

\begin{table*}[t]
\centering
\caption{IntelLabs peer-review detection on the pre-ChatGPT withheld test set: TPR (\%) at different FPR thresholds per source LLM. We compare Anchor, the specialized detector from \citet{yu2025peerreview}, against Vanilla, our fully fine-tuned RoBERTa-base baseline trained on the same calibration subset. Anchor numbers from \citet{yu2025peerreview}.}
\label{tab:intellabs}
\small
\setlength{\tabcolsep}{3pt}
\begin{tabular}{l ccc ccc ccc ccc ccc}
\toprule
& \multicolumn{3}{c}{GPT-4o} & \multicolumn{3}{c}{Gemini-1.5} & \multicolumn{3}{c}{Claude-3.5} & \multicolumn{3}{c}{Llama-3.1} & \multicolumn{3}{c}{Qwen-2.5} \\
\cmidrule(lr){2-4} \cmidrule(lr){5-7} \cmidrule(lr){8-10} \cmidrule(lr){11-13} \cmidrule(lr){14-16}
Method & 0.1\% & 0.5\% & 1.0\% & 0.1\% & 0.5\% & 1.0\% & 0.1\% & 0.5\% & 1.0\% & 0.1\% & 0.5\% & 1.0\% & 0.1\% & 0.5\% & 1.0\% \\
\midrule
Anchor        & 63.5  & 83.7  & 88.8  & 59.7  & 80.3  & 86.5  & 59.6  & 75.8  & 81.8  & 13.6  & 28.5  & 36.8  & 72.5  & 88.4  & 92.1 \\
Vanilla   & \textbf{100.00} & \textbf{100.00} & \textbf{100.00} & \textbf{99.99} & \textbf{100.00} & \textbf{100.00} & \textbf{99.98} & \textbf{100.00} & \textbf{100.00} & \textbf{95.37} & \textbf{98.74} & \textbf{99.39} & \textbf{99.85} & \textbf{99.96} & \textbf{99.99} \\
\bottomrule
\end{tabular}
\end{table*}

\begin{table*}[t]
\centering
\caption{MAGE detection. We compare the long-context Longformer baseline reported in \citet{li2024mage} against Vanilla(large), our fully fine-tuned RoBERTa-large baseline restricted to $512$ tokens. The in-distribution setting uses the standard MAGE split with arbitrary domains and arbitrary generators (Appendix~\ref{app:datasets}); the paraphrasing-attack setting paraphrases every test input with \texttt{gpt-3.5-turbo}, all treated as machine-generated.}
\label{tab:mage}
\small
\setlength{\tabcolsep}{4pt}
\begin{tabular}{lcccc}
\toprule
& \multicolumn{2}{c}{In-distribution} & \multicolumn{2}{c}{Paraphrasing attack} \\
\cmidrule(lr){2-3} \cmidrule(lr){4-5}
Method & AvgRec & AUROC & AvgRec & AUROC \\
\midrule
Longformer (MaxLen$=4096$)   & \textbf{90.5} & \textbf{0.990} & 66.9 & 0.750 \\
Vanilla(large) (MaxLen$=512$)  & 89.2 & 0.982 & \textbf{68.6} & \textbf{0.782} \\
\bottomrule
\end{tabular}
\end{table*}

\subsection{Method: Full Fine-Tuning with RoBERTa}
\label{ssec:fft}

Our baseline detector is deliberately simple: we use RoBERTa~\citep{liu2019roberta} as the backbone and add a single linear classification head over the \texttt{[CLS]} token. The entire backbone is fine-tuned end-to-end, with no frozen layers, auxiliary objectives, or detection-specific architectural components. We train using standard cross-entropy loss against ground-truth labels and optimize with AdamW~\citep{loshchilov2017decoupled}. We refer to this detector as the \emph{vanilla} detector throughout the paper.

We use two backbone sizes, RoBERTa-base ($125$M parameters) and RoBERTa-large ($355$M parameters), and select the size per benchmark to match the model scale of its reported baseline. The corresponding vanilla detectors are denoted \emph{Vanilla(base)} and \emph{Vanilla(large)} when the backbone size needs to be specified. This isolates the effect of full fine-tuning rather than model capacity. For benchmarks with a binary human/machine label, the head outputs two logits; for the three-way setting of FAID~\citep{ta2026faid}, the head outputs three, and we recover a binary score $P(\text{machine}) = 1 - P(\text{human})$ from it when comparing against binary detectors. Remaining training details follow standard practice and are reported in Appendix~\ref{app:training}.

\begin{table*}[!t]
\centering
\caption{FAID three-way classification accuracy. We compare two detectors, LLM-DetectAIve~\citep{abassy2024detectaive} and FAID~\citep{ta2026faid}, against Vanilla and Vanilla $+$ extra, which additionally fine-tunes on each Unseen split's released validation set as a fair analog of FAID's vector-database augmentation. Vanilla reports mean$\pm$std; ``---'' marks splits with no released validation set. LLM-DetectAIve and FAID numbers from \citet{ta2026faid}.}
\label{tab:faid}
\small
\setlength{\tabcolsep}{6pt}
\begin{tabular}{lcccc}
\toprule
Method & FAIDSet (in-dom) & Unseen Domain & Unseen Generator & Unseen Domain$+$Generator \\
\midrule
LLM-DetectAIve        & 94.3            & 52.8            & 75.7            & 62.9 \\
FAID                  & \textbf{95.6}   & 62.8            & \textbf{93.3}   & 66.6 \\
Vanilla           & 93.7$\pm$0.1    & 72.7$\pm$0.8    & 92.7$\pm$1.5    & 67.7$\pm$1.1 \\
Vanilla $+$ extra & ---             & \textbf{91.5$\pm$0.2} & --- & \textbf{88.2$\pm$0.5} \\
\bottomrule
\end{tabular}
% \par\raggedright\footnotesize $^{\dagger}$Unseen Generator has no released validation split; Vanilla is the fair adaptation-equivalent.
\end{table*}

\subsection{Full Fine-Tuning Matches or Exceeds Specialized Detectors}
\label{ssec:exp_baseline}

We now evaluate how the vanilla detector from Section~\ref{ssec:fft} performs in the in-distribution setting, compared with the detector developed in each benchmark. For each of four benchmarks (IntelLabs, FAID, MIRAGE, and MAGE), we reproduce the original paper's in-distribution training and evaluation protocol, i.e., we train on the same data released by the benchmark authors and evaluate on the same test split using the same metrics. This gives a like-for-like comparison between the vanilla detector and the benchmark’s reported detector. 
% Full training details are given in Appendix~\ref{app:training}.

\vspace{0.2em} \noindent\textbf{IntelLabs peer review.}
The IntelLabs benchmark~\citep{yu2025peerreview} introduces Anchor, a detector designed for strict low-FPR operation on academic peer reviews. Our Vanilla(base) baseline is trained on exactly the same calibration subset used to set Anchor's threshold, with two of the five source LLMs (Llama-3.1-70B and Qwen-2.5-72B) unseen by both methods during training (Appendix~\ref{app:datasets}). Table~\ref{tab:intellabs} compares Vanilla(base) with Anchor at the three strict FPR thresholds Anchor is designed for, across the five source LLMs. The gap is large and consistent: Vanilla(base) outperforms Anchor on every LLM at every threshold, with the largest gap at the strictest threshold ($0.1\%$ FPR) on the held-out Llama-3.1-70B.

\paragraph{MAGE.}
MAGE~\citep{li2024mage} is a different kind of benchmark: rather than proposing a new detector, it evaluates several existing detectors on a large in-the-wild corpus and reports a long-context Longformer~\citep{beltagy2020longformer} as the strongest model. Thus, the main comparison is about input length rather than architecture:  the MAGE paper attributes Longformer's advantage to its $4{,}096$-token context window, which is eight times longer than RoBERTa's $512$-token window. 
% We ask whether a $512$-token baseline can match it. 
Table~\ref{tab:mage} compares the $512$-token Vanilla(large) with the $4{,}096$-token Longformer in the two settings. The two models perform similarly in distribution, and Vanilla(large) takes the lead under paraphrasing on both AvgRec and AUROC. This is notable because paraphrasing is exactly the setting where a longer context window is expected to help. Overall, the advantage attributed to long-context modeling on MAGE shrinks substantially when compared against a strong $512$-token baseline, and even reverses under paraphrasing. Additional results on a further out-of-distribution setting are reported in Appendix~\ref{app:c1_tables}.

\vspace{0.2em} \noindent\textbf{FAID.}
Table~\ref{tab:faid} compares Vanilla(base) with FAID's own detector on FAID's three-way classification~\citep{ta2026faid} across four splits: in-domain and three unseen-domain settings. Even without FAID's vector-database augmentation, Vanilla(base) is essentially tied with FAID on the in-domain split and outperforms it on both unseen-domain splits, with the larger margin on Unseen Domain. The released validation sets also allow a small amount of additional fine-tuning per split, which is the fair analogue of FAID's vector-database augmentation; using this additional tuning gives the extended model, Vanilla(base) $+$ extra, which further improves Vanilla(base)'s accuracy.

\vspace{0.2em} \noindent\textbf{MIRAGE.}
Table~\ref{tab:mirage} in Appendix~\ref{app:c1_tables} compares our vanilla detectors with DetectAnyLLM on MIRAGE~\citep{fu2025detectanyllm}. All models are trained on the same $500$ paired GPT-3.5 polish examples used by DetectAnyLLM. At the same model scale, Vanilla(base) matches DetectAnyLLM in AUROC and falls slightly behind on the other metrics; scaling to Vanilla(large) surpasses DetectAnyLLM on all four metrics. We note that the MIRAGE paper~\citep{fu2025detectanyllm} also includes a RoBERTa baseline, but it is not fine-tuned: it corresponds to zero-shot inference with the OpenAI 2019 GPT-2 detector~\citep{solaiman2019release}, precisely the un-fine-tuned baseline model whose weakness motivates this study.

% \paragraph{FAID} 
% The same pattern holds on two further benchmarks. On FAID's three-way classification, a fully fine-tuned RoBERTa-base reproduces the in-domain accuracy of FAID's own detector (within $1.9$ points) and exceeds it on the two unseen-domain splits, by as much as $+9.9$ accuracy points on the Unseen Domain split. Allowing a small amount of additional fine-tuning on each split's released validation set, which is the fair analogue of FAID's vector-database augmentation, raises this further. 

% \paragraph{MIRAGE}
% On MIRAGE, a RoBERTa-base trained on exactly the 500 paired GPT-3.5 polish examples used by DetectAnyLLM matches its $0.934$ average AUROC across the six task subsets at the same model scale; scaling up to RoBERTa-large further reaches $0.963$. The near-chance RoBERTa numbers reported in the MIRAGE paper are not a fine-tuned baseline at all. They correspond to zero-shot inference with the OpenAI 2019 GPT-2 detector~\citep{solaiman2019release}, precisely the un-fine-tuned reference point whose weakness motivates this study. Full per-split results for both benchmarks are given in Appendix~\ref{app:c1_tables} (Tables~\ref{tab:faid} and~\ref{tab:mirage}).

\section{The Distribution-Shift Problem That Remains}
\label{sec:shift}

\begin{table*}[t]
\centering
\caption{Single-source cross-benchmark transfer. Each row is a vanilla detector trained on one benchmark; each column block is a held-out test benchmark, with AUROC and TPR at $1\%$ FPR reported side-by-side. Diagonal cells (in-distribution) shaded. The baseline \texttt{OpenAI 2019}~\citep{solaiman2019release} is included for comparison.}
\vspace{-0.5em}
\label{tab:cross_benchmark_single}
\small
\setlength{\tabcolsep}{2.5pt}
\begin{tabular}{l cc cc cc cc}
\toprule
& \multicolumn{2}{c}{IntelLabs} & \multicolumn{2}{c}{MAGE} & \multicolumn{2}{c}{FAID} & \multicolumn{2}{c}{MIRAGE} \\
\cmidrule(lr){2-3} \cmidrule(lr){4-5} \cmidrule(lr){6-7} \cmidrule(lr){8-9}
Vanilla trained on & AUROC & TPR & AUROC & TPR & AUROC & TPR & AUROC & TPR \\
\midrule
IntelLabs (base) & \cellcolor{gray!15}1.000 & \cellcolor{gray!15}0.999 & 0.533 & 0.034 & 0.785 & \textbf{0.340} & 0.752 & 0.262 \\
MAGE (large)     & 0.931 & 0.529 & \cellcolor{gray!15}0.982 & \cellcolor{gray!15}0.801 & 0.718 & 0.038 & 0.787 & 0.242 \\
FAID (base)      & 0.949 & 0.832 & 0.551 & 0.035 & \cellcolor{gray!15}0.997 & \cellcolor{gray!15}0.924 & \textbf{0.828} & \textbf{0.375} \\
MIRAGE (large)   & \textbf{0.995} & \textbf{0.891} & 0.605 & 0.119 & \textbf{0.920} & 0.042 & \cellcolor{gray!15}0.976 & \cellcolor{gray!15}0.784 \\
\midrule
OpenAI 2019      & 0.605 & 0.113 & \textbf{0.725} & \textbf{0.197} & 0.459 & 0.011 & 0.511 & 0.029 \\
\bottomrule
\end{tabular}
\end{table*}

% \begin{table*}[t]
% \centering
% \caption{Single-source cross-benchmark transfer, measured by TPR at $1\%$ FPR. Each row is a vanilla detector trained on one benchmark, and each column is a test benchmark. Diagonal cells are in-distribution evaluations and are shaded; off-diagonal cells are held-out cross-benchmark evaluations. The \texttt{roberta-base-openai-detector}~\citep{solaiman2019release} reference is included for comparison.}
% % All evaluations use $\text{max\_len}{=}512$. Best off-diagonal per column in \textbf{bold}.
% \label{tab:cross_benchmark_single}
% \small
% \setlength{\tabcolsep}{3pt}
% \begin{tabular}{lcccc}
% \toprule
% Vanilla trained on $\backslash$ Test & IntelLabs & MAGE & FAID & MIRAGE \\
% \midrule
% IntelLabs (base)  & \cellcolor{gray!15}0.999 & 0.034 & 0.340 & 0.262 \\
% MAGE (large)      & 0.529 & \cellcolor{gray!15}0.801 & 0.038 & 0.242 \\
% FAID (base)       & 0.832 & 0.035 & \cellcolor{gray!15}0.924 & 0.375 \\
% MIRAGE (large)    & \textbf{0.891} & \textbf{0.119} & \textbf{0.042} & \cellcolor{gray!15}0.784 \\
% \midrule
% OpenAI 2019       & 0.113 & 0.197 & 0.011 & 0.029 \\
% \bottomrule
% \end{tabular}
% \end{table*}

Section~\ref{sec:baseline} gives a strong but in-distribution result: a plain fully fine-tuned RoBERTa matches or exceeds specialized detectors when the test data is drawn from the same benchmark it was trained on. The more consequential challenge is the distribution shift. In deployment, a detector is trained once but must handle text from unknown topics and unknown generators; in-distribution performance says little about this setting. We therefore ask the central remaining question: how well does this strong baseline perform when the test distribution shifts away from the training distribution, and what can be done when it fails?

% A detector is trained once and then deployed against text whose topic and whose generating model it cannot choose; the in-distribution number says little about that setting. We therefore turn to the question that, in our view, is the one that genuinely remains: how well does this strong baseline hold up when the test distribution moves away from the training distribution, and what can be done when it does not?

\subsection{Distribution Shift Is a Problem In-Distribution Scores Hide}
\label{ssec:shift_problem}

We write $\mathcal{D}_{\text{source}}$ for the labeled distribution used to train a detector and $\mathcal{D}_{\text{target}}$ for the distribution used at evaluation. Section~\ref{sec:baseline} studied the in-distribution case $\mathcal{D}_{\text{source}} = \mathcal{D}_{\text{target}}$; \emph{distribution shift} is the case $\mathcal{D}_{\text{source}} \neq \mathcal{D}_{\text{target}}$. 

We evaluate distribution shift under two protocols with different levels of severity. The first is within-MAGE leave-one-domain-out: we train on $9$ of MAGE's $10$ domains and test on the held-out domain. In this setting, the label space and generator pool are fixed, so only the topic domain changes. This isolates the effect of topic shift. The second is cross-benchmark transfer: we train on one benchmark and evaluate on another, so both the topic domain and the generator distribution may change. For this setting, we also use HC3~\citep{guo2023close}, an early human-versus-ChatGPT corpus covering five English domains, as an additional held-out target benchmark. HC3 is used only as $\mathcal{D}_{\text{target}}$ and is never used to train any detector.

We use within-MAGE to characterize the failure mode, and evaluate the approaches of Section~\ref{ssec:methods} under both protocols.

\begin{figure*}[t]
\centering
\includegraphics[width=\textwidth]{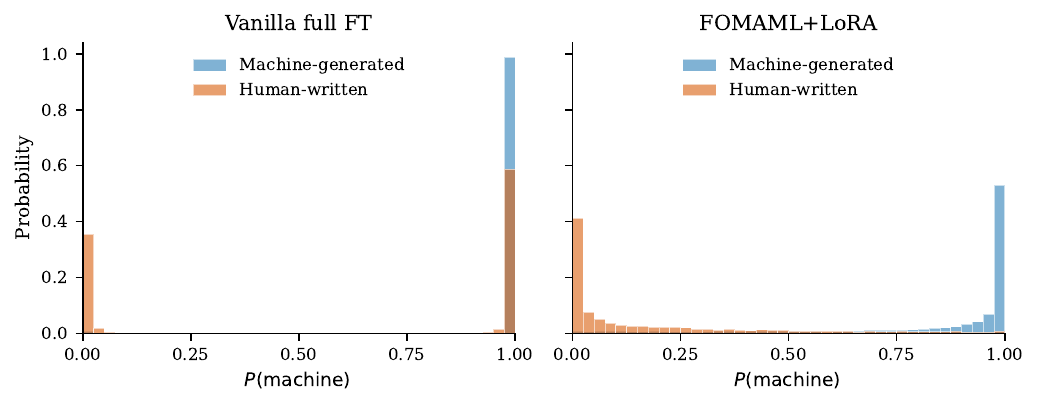}
\vspace{-2em}
\caption{Predicted-probability distributions under within-MAGE leave-one-domain-out, pooled over $10$ held-out domains ($5{,}000$ machine-generated and $5{,}000$ human-written samples per detector). Left: the vanilla detector is sharply concentrated near $0$ and $1$, but many human-written samples (orange) are assigned near-certain machine probabilities, overlapping with machine-generated samples (blue). Right: FOMAML$+$LoRA reduces this failure mode by shifting human-written samples toward $0$ and machine-generated samples toward $1$.}
% \caption{Predicted-probability distribution under within-MAGE leave-one-domain-out, pooled across $10$ held-out domains ($5{,}000$ machine and $5{,}000$ human samples per detector; FOMAML$+$LoRA probabilities averaged over $5$ random support sets). Left: the vanilla detector's predictions are sharply concentrated near $0$ and $1$, and a large fraction of human-written samples (orange) lie in the same high-confidence bin near $1$ as the machine-generated samples (blue), forming the overconfidence described in Section~\ref{ssec:shift_problem}. Right: FOMAML$+$LoRA shifts the human-written distribution toward $0$ and the machine-generated distribution toward $1$, leaving probability mass throughout the interval.}
\label{fig:calibration}
\vspace{-0.5em}
\end{figure*}

\begin{table*}[t]
\centering
\caption{Within-MAGE leave-one-domain-out. Each detector is evaluated on a held-out MAGE domain after training on the other nine. We compare three detectors: Vanilla (our fully fine-tuned RoBERTa-large, no adaptation), FOMAML$+$LoRA ($K{=}10$-shot domain adaptation on the held-out domain), and Ensemble (the per-sample confidence-weighted combination of the two, Eq.~\ref{eq:cwe}). AUROC and TPR at $1\%$ FPR are reported per held-out domain. Within each metric group, best per row in \textbf{bold}.}
\label{tab:c3_within_mage}
\small
\setlength{\tabcolsep}{3pt}
\begin{tabular}{lcccccc}
\toprule
& \multicolumn{2}{c}{Vanilla} & \multicolumn{2}{c}{FOMAML$+$LoRA} & \multicolumn{2}{c}{Ensemble} \\
\cmidrule(lr){2-3} \cmidrule(lr){4-5} \cmidrule(lr){6-7}
Held-out domain & AUROC & TPR & AUROC & TPR & AUROC & TPR \\
\midrule
cmv      & 0.987 & 0.866          & \textbf{0.992} & 0.870 & 0.989          & \textbf{0.888} \\
eli5     & 0.957 & \textbf{0.662} & 0.936          & 0.484 & \textbf{0.958} & 0.636 \\
hswag    & 0.930 & 0.602          & 0.870          & 0.584 & \textbf{0.932} & \textbf{0.660} \\
roct     & 0.865 & 0.374          & 0.867          & 0.446 & \textbf{0.907} & \textbf{0.532} \\
sci\_gen & 0.953 & 0.720          & \textbf{0.969} & 0.694 & 0.957          & \textbf{0.768} \\
squad    & 0.975 & 0.692          & 0.961          & 0.720 & \textbf{0.979} & \textbf{0.788} \\
tldr     & 0.933 & 0.626          & \textbf{0.962} & 0.718 & 0.950          & \textbf{0.750} \\
wp       & 0.971 & 0.552          & \textbf{0.982} & 0.738 & 0.980          & \textbf{0.832} \\
xsum     & 0.825 & 0.274          & \textbf{0.897} & \textbf{0.448} & 0.868 & 0.420 \\
yelp     & 0.945 & 0.508          & 0.944          & 0.548 & \textbf{0.950} & \textbf{0.624} \\
\midrule
\emph{Average} & 0.934 & 0.588 & 0.938 & 0.625 & \textbf{0.947} & \textbf{0.690} \\
\bottomrule
\end{tabular}
\end{table*}

Both distribution-shift protocols show degradation, though with different levels of severity. Cross-benchmark transfer fails most sharply: Table~\ref{tab:cross_benchmark_single} evaluates each single-source detector across different test benchmarks. The diagonal entries give in-distribution performance, while the off-diagonal entries correspond to held-out test benchmarks. At $1\%$ FPR, the off-diagonal entries are far below the diagonal, with several near zero. Within-MAGE leave-one-domain-out is milder but still substantial, with the vanilla detector falling well short of its in-distribution performance across all $10$ held-out rounds (Table~\ref{tab:c3_within_mage}, vanilla column). 

\vspace{0.2em} \noindent\textbf{Distribution shift persists despite more source data.}
Simply pooling benchmarks does not fully solve the problem. Under a leave-one-benchmark-out protocol, pooling several source benchmarks improves over un-fine-tuned references, but the held-out targets still remain far below their in-distribution performance (Table~\ref{tab:multisource}, Appendix~\ref{app:c2c3_tables}). Since no training set can cover every domain a detector may encounter in deployment, adding more source data alone is not enough to overcome distribution shift.

% Pooling several benchmarks under a leave-one-benchmark-out protocol improves on un-fine-tuned references but still leaves held-out targets far below their in-distribution level (Table~\ref{tab:multisource}, Appendix~\ref{app:c2c3_tables}). Since no training set can enumerate every domain a detector will meet at deployment, accumulating more source data is not, on its own, a route out of the problem.

\vspace{0.2em} \noindent\textbf{An overconfidence that AUROC hides.}
Per-domain AUROC under within-MAGE leave-one-domain-out varies widely across the $10$ rounds (Table~\ref{tab:c3_within_mage}), but the more telling column is the TPR at $1\%$ FPR. The drop from in-distribution performance is much larger at this operating point, so AUROC alone would make the degradation look much less severe. The reason becomes clear from the predicted-probability distributions in Figure~\ref{fig:calibration} (Left). Pooling the $10{,}000$ samples across the $10$ held-out domains, $99.4\%$ fall into the highest-confidence bucket, i.e., $p_{\text{vanilla}} \leq 0.05$ or $p_{\text{vanilla}} \geq 0.95$. The detector is almost never uncertain, but its confidence is often misplaced. Within the same high-confidence bucket, the vanilla detector is correct on only $69\%$ of samples: it is not leaving difficult examples in an uncertain middle range, but assigning high-confidence predictions to many errors. The class asymmetry is especially severe. Among the $5{,}000$ human-written samples, $3{,}019$ ($60.4\%$) receive a near-certain machine prediction ($p_{\text{vanilla}} \geq 0.95$). Thus, what appears under AUROC to be a strong detector becomes, at a usable threshold, a detector that confidently labels a majority of human-written text from unseen domains as machine-generated.

\subsection{Domain Adaptation Partially Improves Performance}
\label{ssec:methods}

% \paragraph{Setup.}
Because adding more source data does not close the gap, we turn instead to domain adaptation with two simple approaches. The first is $K$-shot domain adaptation~\citep{finn2017maml, nichol2018firstorder, hu2022lora}: given a small labeled set from the target domain, we adapt the detector with a few gradient steps before evaluation. The second is a per-sample confidence-weighted ensemble, which combines the vanilla detector with the adapted detector.

\vspace{0.2em} \noindent\textbf{Model-Agnostic Meta-Learning (MAML).}
To make $K$-shot adaptation efficient for a fully fine-tuned RoBERTa detector, we implement it as first-order MAML over LoRA adapters (FOMAML$+$LoRA)~\citep{nichol2018firstorder, hu2022lora}. FOMAML meta-trains an initialization that can adapt quickly from a small labeled set. Unlike full MAML, it uses the gradient computed after adaptation as the meta-gradient and avoids second-order Hessian-vector products, making meta-training substantially cheaper. LoRA further improves efficiency by updating only a small set of trainable parameters inserted into the backbone, while keeping the pre-trained weights frozen.

In our setting, each source domain is treated as a task. During meta-training, the LoRA adapters are optimized so that a few gradient steps on $K$ labeled examples per class from any source domain produce a good detector for that domain. At test time, we apply the same adaptation procedure using $K=10$ labeled examples per class from the target domain, starting from the meta-learned initialization. Full equations and hyperparameters are given in Appendix~\ref{app:training}.

% The first is $K$-shot adaptation with first-order MAML over LoRA adapters (FOMAML$+$LoRA)~\citep{nichol2018firstorder, hu2022lora}. FOMAML is a meta-learning procedure that explicitly trains a model initialization to adapt quickly from a small labeled set, while LoRA inserts a small number of trainable parameters into the backbone and keeps the pre-trained weights frozen. In our context, we treat each source domain as a task: at meta-training time, the LoRA adapters are optimized so that a few gradient steps on a $K$-shot support set from any source domain produces a good detector for that domain. At test time, we apply the same few gradient steps using $K = 10$ labeled examples per class from the target domain, adapting from the meta-learned initialization. The second is a per-sample confidence-weighted ensemble of the vanilla and adapted detectors. Full equations and hyperparameters are given in Appendix~\ref{app:training}.

\begin{table*}[!ht]
\centering
\caption{MAGE$\to$HC3 cross-benchmark transfer. The same three detectors as Table~\ref{tab:c3_within_mage}, evaluated under cross-benchmark distribution shift. Vanilla is trained on MAGE and applied zero-shot on each HC3 domain; FOMAML$+$LoRA performs $K{=}10$-shot adaptation per HC3 domain (averaged over $5$ random support sets); Ensemble combines them per Eq.~\ref{eq:cwe}. AUROC and TPR at $1\%$ FPR are reported per HC3 domain. Within each metric group, best per row in \textbf{bold}.}
\label{tab:c3_hc3}
\small
\setlength{\tabcolsep}{3pt}
\begin{tabular}{lcccccc}
\toprule
& \multicolumn{2}{c}{Vanilla} & \multicolumn{2}{c}{FOMAML$+$LoRA} & \multicolumn{2}{c}{Ensemble} \\
\cmidrule(lr){2-3} \cmidrule(lr){4-5} \cmidrule(lr){6-7}
HC3 domain   & AUROC & TPR & AUROC & TPR & AUROC & TPR \\
\midrule
finance      & \textbf{0.997} & \textbf{0.972} & 0.998          & 0.922          & 0.997          & 0.972          \\
medicine     & 0.976          & 0.820          & \textbf{0.991} & 0.813          & 0.979          & \textbf{0.880} \\
open\_qa     & 0.977          & 0.888          & 0.964          & 0.642          & \textbf{0.978} & \textbf{0.894} \\
reddit\_eli5 & 0.997          & 0.924          & 0.998          & 0.966          & 0.997          & \textbf{0.968} \\
wiki\_csai   & 0.992          & 0.875          & \textbf{0.998} & \textbf{0.958} & 0.995          & 0.893          \\
\midrule
\emph{Average} & 0.988 & 0.896 & 0.989 & 0.860 & \textbf{0.989} & \textbf{0.921} \\
\bottomrule
\end{tabular}
\end{table*}

\vspace{0.2em} \noindent\textbf{Ensemble Approach.}
The two detectors have very different prediction profiles (Figure~\ref{fig:calibration}): the vanilla detector concentrates almost all probability mass near $0$ or $1$, while FOMAML$+$LoRA spreads its predictions across the interval. This contrast motivates using the vanilla detector itself as the confidence signal: distance from $0.5$ is a meaningful indicator of (over)confidence for the vanilla detector but not for the adapted one. We therefore combine the two detectors as
\begin{equation}
\hat{p}(x) = w(x)\,p_{\text{vanilla}}(x) + \bigl(1 - w(x)\bigr)\,p_{\text{FOMAML}}(x),
\label{eq:cwe}
\end{equation}
% with $w(x) = 2\,\lvert p_{\text{vanilla}}(x) - 0.5\rvert \in [0,1]$: when the vanilla detector is near-certain ($p_{\text{vanilla}} \to 0$ or $1$), $w(x) \to 1$ and the ensemble follows it; when it is undecided ($p_{\text{vanilla}} \to 0.5$), $w(x) \to 0$ and the ensemble defers to the adapted detector. There are no learned components and no hyperparameters; the only inference-time cost is a second forward pass.
where the weight function $w(x) \in [0,1]$ can be chosen to control how much the ensemble trusts the vanilla detector on each input. We considered several simple choices, including fixed linear weights $\alpha\,p_{\text{vanilla}}(x) + (1-\alpha)\,p_{\text{FOMAML}}(x)$ with $\alpha \in \{0.1,0.2,\ldots,0.9\}$, as well as confidence-based variants that use either detector as the confidence source. The best-performing choice among these strategies is
\begin{equation*}
w(x) = 2\,\lvert p_{\text{vanilla}}(x) - 0.5\rvert \in [0,1].
\end{equation*}
% With this choice, the ensemble follows the vanilla detector when it is near-certain ($p_{\text{vanilla}}(x) \to 0$ or $1$), so $w(x) \to 1$, and defers to the adapted detector when the vanilla detector is uncertain ($p_{\text{vanilla}}(x) \to 0.5$), so $w(x) \to 0$. 
Fixed linear combinations never exceed the better single detector by more than a small margin (Table~\ref{tab:c3_alpha_sweep}), and using FOMAML$+$LoRA as the confidence source leads to substantially worse performance at $1\%$ FPR (Table~\ref{tab:c3_weight_source}). We therefore report Eq.~\ref{eq:cwe} with the vanilla-confidence weight as the empirical winner among the ensemble strategies we tried. 
% The only additional inference-time cost is a second forward pass.

\vspace{0.2em} \noindent\textbf{Both approaches improve performance under distribution shift.}
We evaluate the methods under the within-MAGE leave-one-domain-out setting of Section~\ref{ssec:shift_problem}: detectors are trained on $9$ source domains and tested on the held-out domain. Table~\ref{tab:c3_within_mage} compares the vanilla detector, FOMAML$+$LoRA adaptation, and the confidence-weighted ensemble built on top of the adapted detector across the $10$ held-out MAGE domains. FOMAML$+$LoRA improves over the vanilla baseline at $1\%$ FPR, and the ensemble further improves performance by combining the vanilla and adapted detectors. The gains are modest in AUROC but substantial at the low-FPR operating point, where the distribution-shift failure is most visible. The ensemble improves over vanilla on all $10$ held-out domains, with the largest gains on the domains where vanilla performs worst.

% We evaluate under the within-MAGE leave-one-domain-out setting of Section~\ref{ssec:shift_problem}: detectors are trained on $9$ source domains and tested on the held-out $10$th. Table~\ref{tab:c3_within_mage} compares the vanilla detector, FOMAML$+$LoRA, and the ensemble across the $10$ held-out MAGE domains under leave-one-domain-out. Both methods improve over the vanilla baseline at $1\%$ FPR; the gains are modest in AUROC and substantial at the low-FPR operating point. The ensemble improves over vanilla on all $10$ held-out domains, and the largest gains appear on the domains where vanilla was weakest. That two lightweight approaches, neither introducing a new architecture, both move the low-FPR number in the same direction confirms what Section~\ref{ssec:shift_problem} predicted: the overconfidence is reachable.

% \paragraph{Alternative ensemble strategies.}
% We considered several simple ensemble strategies. Fixed linear combinations 
% $\alpha\,p_{\text{vanilla}} + (1-\alpha)\,p_{\text{FOMAML}}$ swept over 
% $\alpha \in \{0.1, 0.2, \ldots, 0.9\}$ never exceed the better single 
% detector by more than a small margin (Table~\ref{tab:c3_alpha_sweep}). 
% A symmetric variant of Eq.~\ref{eq:cwe} where the two detectors swap roles, with FOMAML$+$LoRA acting as both the dominant predictor and the confidence source, performs substantially worse at $1\%$ FPR (Table~\ref{tab:c3_weight_source}). We report the confidence-weighted ensemble as the empirical winner among the strategies 
% we tried.

\vspace{0.2em} \noindent\textbf{Improvement extends to cross-benchmark transfer.}
The within-MAGE setting keeps the generator pool fixed and changes only the topic domain. We now test a harder cross-benchmark transfer setting, where both the topic domain and the generator distribution change. Specifically, we train the vanilla detector on MAGE and evaluate it on the five English HC3 domains~\citep{guo2023close}: finance, medicine, open-domain QA, Reddit, and Wikipedia. The vanilla detector is applied zero-shot, while FOMAML$+$LoRA performs $K{=}10$-shot adaptation separately for each HC3 domain.
Table~\ref{tab:c3_hc3} reports per-domain AUROC and TPR at $1\%$ FPR for the vanilla detector, FOMAML$+$LoRA, and the confidence-weighted ensemble. The ensemble again improves over the vanilla baseline at $1\%$ FPR, with the largest gain on the medicine domain. The adaptation is also insensitive to the number of labeled target examples: varying $K$ over $\{5, 10, 20, 50\}$ leaves both AUROC and TPR at $1\%$ FPR essentially unchanged (Table~\ref{tab:k_ablation}, Appendix~\ref{app:c2c3_tables}). This suggests that the gain mainly comes from the meta-learned initialization, rather than from the particular labeled examples used for adaptation.

\section{Conclusion}
\label{sec:conclusion}

We revisited AI-generated text detection from a baseline-first perspective, and the first lesson is both simple and consequential: strong general-purpose methods can outperform specialized hand-engineered designs. A plain, fully fine-tuned RoBERTa matches or exceeds the specialized detectors that recent benchmarks are built around, and on the in-the-wild MAGE benchmark, it matches the strongest reported detector once input length is controlled for. This means that a strong and valid baseline is not secondary, but a prerequisite for meaningful innovation. When progress is measured against a zero-shot, un-fine-tuned reference point, that progress has likely been overstated.

% We revisited AI-generated text detection from a baseline-first perspective, and the study yields two findings that pull in different directions. The first is a familiar lesson: general methods tend to outperform hand-engineered ones. A plain, fully fine-tuned RoBERTa matches or exceeds the specialized detectors that recent benchmarks are built around, and on the in-the-wild MAGE benchmark it matches the strongest reported detector once input length is controlled for. To the extent that progress has been measured against a zero-shot, un-fine-tuned reference point, that progress has been overstated.

At the same time, a strong baseline does not make the problem solved. It degrades sharply once the test distribution moves outside the training distribution, and simply accumulating more source data does not close the gap. What helps is domain adaptation: $K$-shot meta-learning with FOMAML$+$LoRA improves performance from only a small labeled target set, and a per-sample confidence-weighted ensemble of the adapted and vanilla detectors provides further gains. This suggests that the remaining bottleneck is not only better detector design, but better evaluation: without testing distribution shift against strong baselines, it is hard to tell whether a method has solved the real deployment problem or only improved an easier in-distribution one.

% The second finding qualifies the first. The strong baseline degrades sharply once the test distribution moves outside the training distribution, and accumulating more source data does not close the gap. What helps is domain adaptation: $K$-shot meta-learning with FOMAML$+$LoRA recovers part of the lost performance from only a small labeled support set, and a per-sample confidence-weighted ensemble of the adapted and vanilla detectors further closes the low-FPR gap. The shape of the degradation points to a blind spot in evaluation: a detector with $0.934$ average AUROC can still place a majority of human-written text in an overconfidence that threshold-independent metrics do not surface. That two lightweight approaches partially close the gap suggests the tail is reachable rather than fundamental, and that the limitation lies in how detectors have been evaluated as much as in the detectors themselves.

Taken together, these findings suggest that progress in AI-generated text detection should be measured against strong fine-tuned baselines, not weak off-the-shelf references. Such baselines do not diminish innovation; they make clear where innovation is actually needed: improving robustness when the distribution changes and ensuring reliable behavior at the low false-positive rates required in deployment.

\newpage
\section*{Limitations}

\paragraph{Input-length truncation.} Our detectors truncate inputs to $512$ tokens. On the benchmarks we evaluate this does not appear to cost detection accuracy. Under the controlled $512$-token comparison on MAGE, the vanilla detector is competitive with a long-context detector even under paraphrasing attack (Section~\ref{ssec:exp_baseline}). Truncation does, however, leave an adversarial surface our experiments do not cover: an input whose first $512$ tokens are human-written, with machine-generated content placed only afterward, would be judged on its human prefix alone and could evade detection. A detector intended to be robust to this would need to process the full input.

\paragraph{Binary formulation of the ensemble.} The confidence-weighted ensemble is defined for binary detection: its weight $w(x) = 2\,|p_{\text{vanilla}}(x) - 0.5|$ is built around the single decision boundary at $0.5$. It does not directly extend to multi-class formulations such as the three-way human/machine/mixed setting of FAID, where no single such boundary exists. A multi-class version of the confidence weighting is left to future work.

\paragraph{Deployment cost and supervision.} Both proposed methods carry deployment assumptions. The ensemble runs two detectors, and so requires a second forward pass at inference. The $K$-shot adaptation assumes a small set of labeled examples from the target distribution; while $K=10$ per class suffices in our experiments, settings with no labeled target data fall outside what it addresses.

\paragraph{The ensemble's gain is conditional on the failure mode being present.} The confidence-weighted ensemble we report is the empirical winner among the simple strategies we tried in the settings of Section~\ref{ssec:methods}, where the vanilla detector exhibits a large high-confidence false-positive tail. We have not established that this ranking holds in regimes where the vanilla detector is already well-behaved on the negative class. The ensemble should be read as an empirical observation tied to the failure mode characterized in Section~\ref{ssec:shift_problem}, not as a generic detector-combination method.

\bibliography{custom}

\newpage

\appendix

\clearpage

\section{Dataset Details}
\label{app:datasets}

This appendix expands the dataset overview of Section~\ref{sec:baseline} and Table~\ref{tab:datasets}, describing the domains, generators, and train/test protocols of each benchmark.

\paragraph{IntelLabs peer review.} The IntelLabs benchmark~\citep{yu2025peerreview} consists of academic peer reviews from two venues, ICLR and NeurIPS. Human-written reviews are treated as reliably human only up to 2022; from 2023 onward they may contain AI-assisted content, so we restrict all experiments to the pre-ChatGPT portion. The dataset provides three paper-level splits that are disjoint by \texttt{unique\_paper\_id} (verified to have zero overlap): a \emph{calibration} split (ICLR and NeurIPS 2021--2022, $2{,}000$ unique papers), a \emph{test} split (the pre-ChatGPT withheld set, $5{,}406$ unique papers across multiple years), and an \emph{extended} split. Machine-generated reviews are produced by five source LLMs: GPT-4o, Gemini-1.5-Pro, Claude-Sonnet-3.5, Llama-3.1-70B, and Qwen-2.5-72B. We train only on the calibration split, dividing it $90/10$ into training and validation via \texttt{GroupShuffleSplit} grouped on \texttt{unique\_paper\_id}; grouping by paper, rather than by individual review, prevents multiple reviews of the same paper from being split across training and validation. Anchor itself has no trainable parameters: it learns only a per-LLM threshold on the calibration set, using reference reviews generated from GPT-4o, Gemini-1.5-Pro, and Claude-Sonnet-3.5. Our Vanilla(base) baseline is trained on the same calibration subset (ICLR reviews from these three LLMs), so that the comparison in Section~\ref{ssec:exp_baseline} holds Llama-3.1-70B and Qwen-2.5-72B out of training for both methods.

\paragraph{MAGE.} MAGE~\citep{li2024mage} is a $437$K-example corpus spanning ten topic domains (\texttt{cmv}, \texttt{eli5}, \texttt{hswag}, \texttt{roct}, \texttt{sci\_gen}, \texttt{squad}, \texttt{tldr}, \texttt{wp}, \texttt{xsum}, and \texttt{yelp}), with machine text from twenty-seven generators. The original paper defines eight testbeds of increasing distributional difficulty: Testbeds 1--4 are in-distribution (training and test drawn from the same sources, varying whether the domain and the generator set are fixed or arbitrary), and Testbeds 5--8 are out-of-distribution (test data from unseen models, unseen domains, or both). We evaluate on Testbed~4 (arbitrary domains and arbitrary models, the most comprehensive in-distribution setting), Testbed~7 (four held-out datasets, namely CNN/DailyMail, DialogSum, PubMedQA, and IMDb, with GPT-4-generated machine text), and Testbed~8 (Testbed~7 with every text additionally paraphrased by \texttt{gpt-3.5-turbo}, all treated as machine-generated). We focus on these three because they are released as ready-to-use splits, they share a single training set so that one trained model evaluates all three, and they cover the most informative in-distribution, out-of-distribution, and adversarial regimes; Testbeds 1--3 show little separation between methods, and Testbeds 5--6 require leave-one-out cross-validation with many separately trained classifiers.

For the within-MAGE leave-one-domain-out experiments (Section~\ref{ssec:methods}), each domain is split $80/20$ with a fixed seed; the evaluation portion of each class is capped at $\min(500, 20\%)$ examples. In each round, the held-out domain contributes only its evaluation portion as the test set, while the remaining nine domains contribute only their training portions to the pooled training set. The split of each domain is identical across all rounds, so no test example of a held-out domain ever appears in another round's training pool.

\paragraph{FAID.} FAID~\citep{ta2026faid} is an academic-text benchmark with a three-way label space: human-written, fully machine-generated, and human-machine collaborative. Its training set, FAIDSet/train, covers two academic domains, student theses and paper abstracts, with machine text from four LLM families (GPT, Gemini, DeepSeek, and Llama). FAID defines four evaluation splits according to whether the topic domain and the generator are seen during training: \emph{in-domain} (same domains and generators as training), \emph{Unseen Domain} (IELTS essays, with the four training generators), \emph{Unseen Generator} (academic text, with three unseen generators: Qwen, Mistral, and Gemma), and \emph{Unseen Domain $+$ Generator} (IELTS essays with the three unseen generators). The Unseen Generator split is released with a test set only and no validation split; consequently, the additional-fine-tuning variant we report on the other unseen splits cannot be applied to it. We use the English portion of FAID only.

\paragraph{MIRAGE.} MIRAGE~\citep{fu2025detectanyllm} is sampled from five text domains, with machine text from seventeen LLMs. Its detector, DetectAnyLLM, is trained on $500$ paired GPT-3.5 polish examples, which we use as the training data for our vanilla detectors to enable a fair comparison. The test set comprises six subsets, formed by crossing two input settings with three revision tasks. The two input settings are Shared-Input Generation (SIG), in which human and machine texts share the same input prompt, and Disjoint-Input Generation (DIG), in which they do not. The three revision tasks are Generate, Polish (small-scale stylistic revision of existing human text), and Rewrite. Because the SIG subsets are class-imbalanced, we additionally report balanced accuracy and MCC alongside AUROC for MIRAGE.

\paragraph{HC3.} HC3~\citep{guo2023close} is an early human-versus-ChatGPT comparison corpus of question-answering text, with ChatGPT as the single generator. We use its five English domains: finance, medicine, open\_qa, reddit\_eli5, and wiki\_csai. HC3 is not used to train any source detector; it serves only as a target distribution for the MAGE$\to$HC3 cross-benchmark experiments, where the vanilla detector is evaluated zero-shot and FOMAML$+$LoRA performs $K$-shot adaptation using a small support set drawn from each HC3 domain.

\section{Training Details}
\label{app:training}

This appendix gives the training and optimization details omitted from Sections~\ref{sec:baseline} and~\ref{sec:shift}.

\paragraph{Vanilla detector.}
All vanilla detectors are fine-tuned end-to-end with AdamW, a learning rate of $2 \times 10^{-5}$ for Vanilla(base) and $1 \times 10^{-5}$ for Vanilla(large), a batch size of $32$, $3$ epochs, a warmup ratio of $0.06$, and a maximum input length of $512$ tokens; inputs longer than $512$ tokens are truncated. The same configuration is used for every benchmark in Section~\ref{sec:baseline} and for the single-source and pooled detectors in Section~\ref{sec:shift}; only the backbone size varies, selected per benchmark as described in Section~\ref{ssec:fft}.

\paragraph{FOMAML$+$LoRA.}
The LoRA adapters use rank $r = 16$, scaling factor $\alpha = 32$, and dropout $0$, and are inserted into the query and value projections of every self-attention layer of Vanilla(large); the backbone is frozen. This yields approximately $2.6$M trainable parameters, about $0.7\%$ of the full model.

\begin{table*}[!ht]
\centering
\caption{MIRAGE detection, averaged across six task subsets (DIG/SIG $\times$ generate/polish/rewrite). Our vanilla detectors are fine-tuned on the $500$ paired GPT-3.5 polish examples and report mean$\pm$std over 3 seeds.}
\label{tab:mirage}
\small
\setlength{\tabcolsep}{3pt}
\begin{tabular}{lcccc}
\toprule
Method & AUROC & BAcc & MCC & TPR@5\% \\
\midrule
RoBERTa-base (OpenAI 2019)  & 0.510           & 0.516           & 0.044           & 0.074 \\
RoBERTa-large (OpenAI 2019) & 0.515           & 0.526           & 0.053           & 0.080 \\
DetectAnyLLM               & 0.934           & 0.881           & 0.764           & 0.773 \\
Vanilla(base)                & 0.934$\pm$0.010 & 0.871$\pm$0.010 & 0.746$\pm$0.017 & 0.767$\pm$0.018 \\
Vanilla(large)               & \textbf{0.963$\pm$0.009} & \textbf{0.909$\pm$0.019} & \textbf{0.820$\pm$0.038} & \textbf{0.859$\pm$0.042} \\
\bottomrule
\end{tabular}
\end{table*}

\begin{table*}[!ht]
\centering
\caption{MAGE additional out-of-distribution evaluation on a held-out split (CNN/DailyMail, DialogSum, PubMedQA, IMDb, with GPT-4-generated machine text), changing both topic domain and generator relative to training.}
\label{tab:mage_appendix}
\small
\setlength{\tabcolsep}{6pt}
\begin{tabular}{lcc}
\toprule
Method & AvgRec & AUROC \\
\midrule
Longformer (paper, 4096)   & 75.8          & 0.940 \\
Vanilla(large) (ours, 512)  & \textbf{76.0} & \textbf{0.969} \\
\bottomrule
\end{tabular}
\end{table*}

\begin{table*}[!ht]
\centering
\caption{K-shot sensitivity analysis on MAGE$\to$HC3 cross-benchmark transfer. FOMAML$+$LoRA performance averaged across the five HC3 domains (finance, medicine, open\_qa, reddit\_eli5, wiki\_csai), with $K$ varying over $\{5, 10, 20, 50\}$. Each entry is averaged over 5 random support sets per domain. Variation across $K$ is small in both metrics, supporting the claim that the meta-learned initialization, not the particular support examples, carries the adaptation.}
\label{tab:k_ablation}
\small
\setlength{\tabcolsep}{8pt}
\begin{tabular}{ccc}
\toprule
$K$ & AUROC & TPR @ $1\%$ FPR \\
\midrule
5  & 0.989 & 0.873 \\
10 & 0.989 & 0.864 \\
20 & 0.990 & 0.877 \\
50 & 0.989 & 0.873 \\
\bottomrule
\end{tabular}
\end{table*}

\begin{algorithm}[!ht]
\caption{FOMAML$+$LoRA meta-training and test-time adaptation}
\label{alg:fomaml_lora}
\begin{algorithmic}[1]
\Require Source tasks $\{\mathcal{T}_1, \ldots, \mathcal{T}_T\}$; target task $\mathcal{T}_{\text{target}}$; $\eta_{\text{meta}}, \eta_{\text{inner}}, n, K, M$
\State Initialize $\theta$ (LoRA adapters $+$ head); freeze backbone
\For{$\text{iter} = 1, \ldots, M$} \Comment{Meta-training}
    \State Sample task batch from $\{\mathcal{T}_i\}$; for each $\mathcal{T}_i$ sample $S_i, Q_i$
    \State For each $\mathcal{T}_i$: $\theta_i^{(n)} \gets \textsc{InnerLoop}(\theta, S_i, n)$, $g_i \gets \nabla_{\theta_i^{(n)}} \mathcal{L}_{Q_i}(\theta_i^{(n)})$
    \State $\theta \gets \theta - \eta_{\text{meta}} \cdot \text{Adam}(\tfrac{1}{|\text{batch}|}\sum_i g_i)$
\EndFor
\State \textbf{return} $\theta_{\text{target}}^{(n)} \gets \textsc{InnerLoop}(\theta, S_{\text{target}}, n)$ \Comment{Test-time adaptation on $\mathcal{T}_{\text{target}}$}
\Statex
\Function{InnerLoop}{$\theta_0, S, n$}
    \State $\theta^{(0)} \gets \theta_0$
    \For{$s = 0, \ldots, n-1$}
        \State $\theta^{(s+1)} \gets \theta^{(s)} - \eta_{\text{inner}} \nabla_{\theta} \mathcal{L}_S(\theta^{(s)})$
    \EndFor
    \State \textbf{return} $\theta^{(n)}$
\EndFunction
\end{algorithmic}
\end{algorithm}

Let $\theta$ collect the trainable parameters (the LoRA adapters and the classification head). Each source domain is treated as a task. Algorithm~\ref{alg:fomaml_lora} summarizes the full procedure. At every meta-iteration we sample a batch of tasks; for a task $\mathcal{T}_i$ with a support set $S_i$ and a query set $Q_i$, we run $n$ steps of inner-loop SGD on the support set,
\begin{equation}
\theta_i^{(s+1)} = \theta_i^{(s)} - \eta_{\text{inner}} \nabla_{\theta}
\mathcal{L}_{S_i}\!\left(\theta_i^{(s)}\right),
\label{eq:inner_loop}
\end{equation}
starting from $\theta_i^{(0)} = \theta$ and producing adapted parameters $\theta_i^{(n)}$. First-order MAML then uses the gradient evaluated \emph{at} the adapted parameters as the meta-gradient, applied with Adam~\citep{kingma2014adam}:
\begin{equation}
\theta \leftarrow \theta - \eta_{\text{meta}} \cdot \text{Adam}\!\left(
\frac{1}{T} \sum_{i=1}^{T} \nabla_{\theta_i^{(n)}}
\mathcal{L}_{Q_i}\!\left(\theta_i^{(n)}\right) \right),
\label{eq:meta_update}
\end{equation}
which avoids the second-order Hessian-vector products of full MAML at little cost in performance~\citep{nichol2018firstorder}.

Meta-training runs for $2000$ meta-iterations with an outer learning rate $\eta_{\text{meta}} = 10^{-4}$ (Adam); each task uses $K = 10$ support examples per class, $|Q_i| = 16$ query examples per class, and $n = 3$ inner-loop SGD steps with inner learning rate $\eta_{\text{inner}} = 10^{-3}$. At test time, given an unseen target domain, we draw a support set of $K$ labeled examples per class from its training pool, run the same $n$ inner-loop steps from the meta-learned $\theta$, and evaluate the adapted detector on the target test set. Every reported FOMAML$+$LoRA number is averaged over $5$ independently drawn random support sets to control for support-set variance.

\paragraph{Random seeds.}
The in-distribution benchmark comparisons of Section~\ref{sec:baseline} that report error bars (Tables~\ref{tab:faid} and~\ref{tab:mirage}) are averaged over $3$ random seeds ($42$, $123$, $2024$). The cross-benchmark and within-MAGE adaptation experiments of Section~\ref{sec:shift} use a single fixed seed ($42$), as their compute cost makes repeated runs impractical; tables reporting these results are marked accordingly.

\paragraph{Data leakage control.}
For the pooled cross-benchmark training of Section~\ref{ssec:shift_problem}, we remove any training example sharing a $30$-token n-gram with a held-out test example, so that reported transfer numbers are not inflated by overlap between the pooled sources and the target test set.

\section{Additional Results for Section~\ref{ssec:exp_baseline}}
\label{app:c1_tables}

This appendix reports the full per-benchmark results for MIRAGE (Table~\ref{tab:mirage}) summarized in Section~\ref{ssec:exp_baseline}, together with an additional MAGE out-of-distribution evaluation (Table~\ref{tab:mage_appendix}).

\paragraph{MAGE held-out domains.}
Beyond the in-distribution and paraphrasing-attack settings of Table~\ref{tab:mage}, MAGE provides a further out-of-distribution split with four held-out datasets (CNN/DailyMail, DialogSum, PubMedQA, IMDb) paired with GPT-4-generated machine text, changing both topic domain and generator at once. Table~\ref{tab:mage_appendix} shows a fully fine-tuned Vanilla(large) operating on $512$ tokens matches the long-context Longformer on AvgRec ($76.0$ vs $75.8$) and exceeds it on AUROC ($0.969$ vs $0.940$), consistent with the in-distribution and paraphrasing-attack patterns of Section~\ref{ssec:exp_baseline}.

\section{Additional Results for Section~\ref{ssec:methods}}
\label{app:c2c3_tables}

This appendix reports additional results for Section~\ref{ssec:methods}: the $K$-shot sensitivity analysis on MAGE$\to$HC3 transfer (Table~\ref{tab:k_ablation}), the multi-source pooling under leave-one-benchmark-out (Table~\ref{tab:multisource}), the fixed mixing coefficient versus per-sample weighting comparison (Table~\ref{tab:c3_alpha_sweep}), the comparison between symmetric ensemble designs (Table~\ref{tab:c3_weight_source}), and an oracle analysis of how much low-FPR information remains in the two detector scores (Table~\ref{tab:binned-lr-benchmark}).

For the oracle analysis, we ask how much additional low-FPR signal remains in the two detector scores beyond the deployable confidence-weighted ensemble. By the Neyman--Pearson perspective, the optimal target-aware rule in a given score space ranks examples by the likelihood ratio between the machine and human score distributions. We therefore estimate binned empirical likelihood-ratio rules using target-domain labels, both in the joint score space $(p_v,p_f)$ and in each single-score space. As shown in Table~\ref{tab:binned-lr-benchmark}, the 2-D oracle improves over the ensemble on most domains, while both 1-D oracles remain below the ensemble on average, suggesting that the residual oracle headroom comes from the joint two-score structure rather than from either score alone.

% \begin{table*}[!ht]
% \centering
% \caption{AUROC for the same detectors as Table~\ref{tab:cross_benchmark_single}. Row: vanilla training benchmark; column: held-out test benchmark. Diagonal cells (in-distribution) shaded. Two reference points are included: a non-fine-tuned RoBERTa-base and \texttt{roberta-base-openai-detector}~\citep{solaiman2019release}.}
% \label{tab:cross_benchmark_auroc}
% \small
% \setlength{\tabcolsep}{3pt}
% \begin{tabular}{lcccc}
% \toprule
% RoBERTa trained on $\backslash$ Test & IntelLabs & MAGE & FAID & MIRAGE \\
% \midrule
% IntelLabs (base)   & \cellcolor{gray!15}1.000 & 0.533 & 0.785 & 0.752 \\
% MAGE (large)       & 0.931 & \cellcolor{gray!15}0.982 & 0.718 & 0.787 \\
% FAID (base)        & 0.949 & 0.551 & \cellcolor{gray!15}0.997 & \textbf{0.828} \\
% MIRAGE (large)     & \textbf{0.995} & 0.605 & \textbf{0.920} & \cellcolor{gray!15}0.976 \\
% \midrule
% No fine-tuning     & 0.544 & 0.516 & 0.457 & 0.517 \\
% OpenAI 2019        & 0.605 & \textbf{0.725} & 0.459 & 0.511 \\
% \bottomrule
% \end{tabular}
% \end{table*}

\begin{table*}[!ht]
\centering
\caption{Multi-source pooling under leave-one-benchmark-out evaluation (AUROC). ``4-way Mixed'' merges three non-held-out benchmarks via per-benchmark capping; ``Stratified base/large'' uses six-super-domain stratified sampling. ``Best Single'' is the strongest cross-benchmark detector for that column from Table~\ref{tab:cross_benchmark_single}.}
\label{tab:multisource}
\small
\setlength{\tabcolsep}{4pt}
\begin{tabular}{lcccc}
\toprule
Held-out test & 4-way Mixed & Strat base & Strat large & Best Single \\
\midrule
IntelLabs & 0.968 & 0.970 & \textbf{0.997} & 0.995 \\
MAGE      & 0.592 & 0.565 & 0.610          & \textbf{0.725}$^{\dagger}$ \\
FAID      & 0.764 & 0.779 & 0.828          & \textbf{0.920} \\
MIRAGE    & 0.777 & 0.827 & \textbf{0.868} & 0.828 \\
\bottomrule
\end{tabular}
\par\raggedright\footnotesize $^{\dagger}$Best Single for MAGE is OpenAI 2019 detector (Table~\ref{tab:cross_benchmark_single}), which benefits from GPT-2-era LLM overlap with MAGE training data.
\end{table*}

\begin{table*}[!ht]
\centering
\caption{Fixed mixing coefficient versus the per-sample confidence-weighted ensemble, on within-MAGE leave-one-domain-out (averaged over $10$ held-out domains). Each row mixes the two detectors as $p = \alpha\,p_{\text{vanilla}} + (1-\alpha)\,p_{\text{FOMAML}}$ with a constant $\alpha$. The confidence-weighted ensemble (Eq.~\ref{eq:cwe}) uses the per-sample weight $w(x)$ instead.}
\label{tab:c3_alpha_sweep}
\small
\setlength{\tabcolsep}{6pt}
\begin{tabular}{lcc}
\toprule
Mixing & AUROC & TPR@$1\%$FPR \\
\midrule
$\alpha=0.00$ (FOMAML$+$LoRA)   & 0.938 & 0.625 \\
$\alpha=0.50$                   & 0.941 & 0.629 \\
$\alpha=0.70$                   & 0.941 & 0.631 \\
$\alpha=0.80$                   & 0.942 & 0.632 \\
$\alpha=0.85$                   & 0.942 & 0.635 \\
$\alpha=0.90$                   & 0.942 & 0.636 \\
$\alpha=1.00$ (vanilla) & 0.934 & 0.588 \\
\midrule
Confidence-weighted & \textbf{0.947} & \textbf{0.690} \\
\bottomrule
\end{tabular}
\end{table*}

\begin{table*}[!t]
\centering
\caption{Two symmetric ensemble designs, evaluated on within-MAGE leave-one-domain-out ($10$-domain average). The two designs differ only in which detector acts as the dominant predictor and as the confidence source for the per-sample weight $w$.}
\label{tab:c3_weight_source}
\small
\setlength{\tabcolsep}{8pt}
\begin{tabular}{lcc}
\toprule
Ensemble & AUROC & TPR @ $1\%$ FPR \\
\midrule
$w \cdot p_{\text{vanilla}} + (1-w) \cdot p_{\text{FOMAML}}$, \quad $w = 2\,|p_{\text{vanilla}} - 0.5|$ & \textbf{0.947} & \textbf{0.690} \\
$w \cdot p_{\text{FOMAML}} + (1-w) \cdot p_{\text{vanilla}}$, \quad $w = 2\,|p_{\text{FOMAML}} - 0.5|$ & 0.928 & 0.497 \\
\bottomrule
\end{tabular}
\end{table*}

\begin{table*}[t]
\centering
\caption{\textbf{Binned empirical likelihood-ratio benchmark in the detector-score space.}
We compare the confidence-weighted ensemble (Eq.~\ref{eq:cwe}) with target-aware binned LR oracle diagnostics built from the joint score $Z=(\operatorname{logit}p_v,\operatorname{logit}p_f)$ and from each single score alone. The joint oracle uses $M{=}14$ quantile bins per axis with Laplace smoothing $\lambda{=}0.5$; the 1-D oracles use the analogous one-dimensional binning. TPR is evaluated at the largest empirical operating point with $\mathrm{FPR}\le 1\%$, matching Table~\ref{tab:c3_within_mage}. All binned LR rules use target-domain labels and should therefore be interpreted only as oracle diagnostics.}
\label{tab:binned-lr-benchmark}
\small
\setlength{\tabcolsep}{5pt}
\begin{tabular}{lccccc}
\toprule
Held-out domain
& Ensemble TPR
& 2D LR TPR
& $\Delta\mathrm{TPR}$
& 1D-$p_v$ TPR
& 1D-$p_f$ TPR \\
\midrule
cmv      & 0.888 & 0.942 & $+0.054$ & 0.848 & 0.846 \\
eli5     & 0.636 & 0.728 & $+0.092$ & 0.607 & 0.426 \\
hswag    & 0.660 & 0.702 & $+0.042$ & 0.590 & 0.552 \\
roct     & 0.532 & 0.620 & $+0.088$ & 0.307 & 0.410 \\
sci\_gen & 0.768 & 0.786 & $+0.018$ & 0.708 & 0.670 \\
squad    & 0.788 & 0.870 & $+0.082$ & 0.627 & 0.709 \\
tldr     & 0.750 & 0.824 & $+0.074$ & 0.611 & 0.701 \\
wp       & 0.832 & 0.824 & $-0.008$ & 0.615 & 0.708 \\
xsum     & 0.420 & 0.620 & $+0.200$ & 0.292 & 0.422 \\
yelp     & 0.624 & 0.692 & $+0.068$ & 0.566 & 0.524 \\
\midrule
\textbf{Average}
& \textbf{0.690}
& \textbf{0.761}
& $\mathbf{+0.071}$
& \textbf{0.577}
& \textbf{0.597} \\
\bottomrule
\end{tabular}
\end{table*}

\end{document}